\documentclass[sigconf]{acmart}
\usepackage{subfigure}
 
\usepackage{amsmath,amssymb,amsfonts}
\usepackage{bm}
\usepackage{threeparttable}
\usepackage{multirow}
\usepackage{xspace}

\newcommand{\method}{\textit{SofaNet}\xspace}

\AtBeginDocument{%
  \providecommand\BibTeX{{%
    \normalfont B\kern-0.5em{\scshape i\kern-0.25em b}\kern-0.8em\TeX}}}

\setcopyright{acmcopyright}
\copyrightyear{2023}
\acmYear{2023}
\setcopyright{acmlicensed}\acmConference[WWW '23]{Proceedings of the ACM Web Conference 2023}{April 30-May 4, 2023}{Austin, TX, USA}
\acmBooktitle{Proceedings of the ACM Web Conference 2023 (WWW '23), April 30-May 4, 2023, Austin, TX, USA}
\acmPrice{15.00}
\acmDOI{10.1145/3543507.3583989}
\acmISBN{978-1-4503-9416-1/23/04}

\begin{document}

\title{Cross-center Early Sepsis Recognition by Medical Knowledge Guided Collaborative Learning for Data-scarce Hospitals}

\author{Ruiqing Ding$^{1,2}$, Fangjie Rong$^{4}$, Xiao Han$^{*3}$, Leye Wang\authornote{Corresponding authors}$^{*1,2}$}

\affiliation{
$^{1}$\institution{Key Lab of High Confidence Software Technologies (Peking University), Ministry of Education \country{China}} 
$^{2}$\institution{School of Computer Science, Peking University, Beijing, China}
$^{3}$\institution{School of Information Management and Engineering, Shanghai University of Finance and Economics, Shanghai, China}
$^{4}$\institution{School of Mathematical Science, Peking University, Beijing, China}
}
\email{ruiqingding@stu.pku.edu.cn, 1900010621@pku.edu.cn, xiaohan@mail.shufe.edu.cn, leyewang@pku.edu.cn}

\renewcommand{\authors}{Ruiqing Ding,  Fangjie Rong, Xiao Han and Leye Wang}
\renewcommand{\shortauthors}{Ruiqing Ding,  Fangjie Rong, Xiao Han and Leye Wang}

\begin{abstract}
There are significant regional inequities in health resources around the world.
It has become one of the most focused topics to improve health services for data-scarce hospitals and promote health equity through knowledge sharing among medical institutions.
Because electronic medical records (EMRs) contain sensitive personal information, privacy protection is unavoidable and essential for multi-hospital collaboration.
In this paper, for a common disease in ICU patients, sepsis, we propose a novel cross-center collaborative learning framework guided by medical knowledge, \method, to achieve early recognition of this disease.
The Sepsis-3 guideline, published in 2016, defines that sepsis can be diagnosed by satisfying both suspicion of infection and Sequential Organ Failure Assessment (SOFA) greater than or equal to 2.
Based on this knowledge, \method adopts a multi-channel GRU structure to predict SOFA values of different systems, which can be seen as an auxiliary task to generate better health status representations for sepsis recognition.
Moreover, we only achieve feature distribution alignment in the hidden space during cross-center collaborative learning, which ensures secure and compliant knowledge transfer without raw data exchange.
Extensive experiments on two open clinical datasets, MIMIC-III and Challenge, demonstrate that \method can benefit early sepsis recognition when hospitals only have limited EMRs.

\end{abstract}



\maketitle

\section{Introduction}
Significant disparities in health resources have always existed, not only in developed and developing countries but even in different areas and among different ethnicities within the same country. 
Because of the restriction of health resources and service level, it is overwhelming for medical institutions in less developed areas to early diagnosis and timely clinical management of some noncommunicable diseases, e.g., sepsis, diabetes, and heart diseases. 
Enhancing health services in less developed regions is important to promote health equity.\footnote{https://www.who.int/health-topics/health-equity}
By leveraging web and AI techniques, recent efforts have attempted to connect multiple medical institutions for knowledge sharing to improve the service for data-scarce hospitals \cite{lee2012adapting,gupta2020transfer,kachuee2018ecg,weimann2021transfer}. 
However, medical data involves individuals' private and sensitive information, and thus directly transmitting these datasets will inevitably lead to severe privacy violations \cite{miotto2018deep,price2019privacy}. 
In essence, enhancing health equity for medical institutions lacking data resources remains a critical issue on a global scale.

In this paper, we use \textit{early sepsis recognition} as the representative task to study how to improve health equity for medical institutions without sufficient data, considering that sepsis is one of the most serious medical conditions causing millions of deaths with significant regional disparity.
Sepsis is a life-threatening organ dysfunction resulting from a dysregulated host response to infection \cite{10.1001/jama.2016.0287}.
If not detected early and treated promptly, it can result in septic shock, multiple organ failure, and even death. 
Owing to the complexity and importance of clinical sepsis diagnosis and treatment, there are multiple versions of sepsis consensus definitions and diagnosis guidelines, including Sepsis-1 (1991) \cite{american1992american}, Sepsis-2 (2001) \cite{levy_2001_2003} and Sepsis-3 (2016) \cite{10.1001/jama.2016.0287}.
It was estimated that there were 48.9 million cases and 11 million sepsis-related deaths worldwide in 2017, accounting for almost 20\% of all global deaths  \cite{rudd2020global}.
Moreover, significant regional disparities exist in sepsis incidence and mortality --- approximately 85\% of sepsis cases and sepsis-related deaths occurred in low- and middle-income countries, specifically with the highest burden in sub-Saharan Africa, Oceania, south Asia, east Asia, and southeast Asia \cite{rudd2020global}.
Prior research has suggested that the sepsis mortality rate may increase by 7\% for every one-hour delay in the administration of antibiotic treatment \cite{kumar2006duration}. 
Therefore, early recognition is a crucial first step in the management of sepsis.

Nowadays, machine learning techniques are broadly studied in early sepsis recognition and diagnosis, such as the linear model \cite{shashikumar2017early}, neural network \cite{futoma2017learning}, GBDT \cite{li2020time}, etc.
These methods require a large amount of training data to guarantee performance.
Unfortunately, a prior worldwide data challenge \cite{reyna2019early} has revealed that the early sepsis recognition model learned from a hospital's data may not work well for another hospital.
However, it is unrealistic to have large-scale electronic medical records for every hospital, especially for sparsely populated areas where the admitted patients are limited.
To cope with the limitations of small data, some academic studies have built models with data from multiple hospitals, called multicenter study \cite{ferrer2008improvement,xie2020epidemiology}.
Nevertheless, most multicenter studies do not consider the potential privacy leakage when different centers' patient data are gathered.

In this research, to strengthen the ability of early sepsis recognition for medical institutions without sufficient data, we investigate two possible strategies: (i) incorporating domain knowledge in healthcare model design to relieve the data limitation, and (ii) enabling cross-center collaborations between medical institutions in a privacy-preserving manner. 
Accordingly, we propose a cross-center collaborative learning framework, \method, to realize early sepsis recognition with two main components: (i) \textit{the multi-channel recurrent neural network structure} to predict SOFA (Sequential Organ Failure Assessment) scores of multiple systems which are highly associated with sepsis diagnosis (according to the guidelines of Sepsis-3 \cite{10.1001/jama.2016.0287}), and (ii) \textit{the cross-center feature distribution alignment component} to achieve effective knowledge transfer without raw data sharing. 
Our contributions are as follows:

(i) To the best of our knowledge, this is one of the pioneering studies to design a privacy-preserving cross-center collaboration mechanism for early sepsis recognition by explicitly considering domain knowledge (i.e., multi-system SOFA scores).

(ii) By conducting the transfer experiments on two open clinical datasets, MIMIC-III and Challenge, we have validated taht \method significantly and consistently outperforms the start-of-the-art methods without raw clinical data exchange. 
We release our code at https://doi.org/10.5281/zenodo.7625404.

\section{Related Work}
Machine learning techniques are excellent at analyzing complex signals in data-rich environments which promise the effectiveness of early sepsis recognition.
Most studies are carried out in the ICU \cite{fleuren2020machine,karpatne2018machine}. 
The systematic review and meta-analysis indicate that individual machine learning models can accurately predict the onset of sepsis in advance on retrospective data \cite{li2020time,8624374}. 
The PhysioNet/Computing in Cardiology (CinC) Challenge 2019 focused on this issue and promoted the development of open-source AI algorithms for real-time and early recognition of sepsis \cite{reyna2019early}.
However, there are few studies that concentrate on sepsis recognition without sufficient data and the common method is centralized learning (i.e., put data together to analysis \cite{xie2020epidemiology}).
Recently, transfer learning and multi-task learning are becoming popular to utilize knowledge shared by different datasets or tasks to achieve better model performance \cite{pan2009survey}.
In healthcare, some work focuses on a specific disease to design transfer methods, such as blood pressure \cite{leitner2021personalized}, heart disease \cite{kachuee2018ecg}, Covid-19 \cite{ma2021distilling}, etc.; there are also work focusing on privacy issues and designing corresponding algorithms \cite{ju2020privacy}.

\section{Problem Formulation}

\textbf{Early Sepsis Recognition.}
The objective is to use patients' electronic medical records (EMRs) to predict the risk of sepsis.
Considering the early warning of sepsis is potentially life-saving, we recognize sepsis onset in the next 6 hours with patients' last 6-hour EMRs, including vital variables, laboratory variables and demographic information (details in Appendix).
This setting is consistent with the PhysioNet Computing in Cardiology Challenge 2019 \cite{reyna2019early,goldberger2000physiobank} on \textit{Early Prediction of Sepsis from Clinical Data}\footnote{https://physionet.org/content/challenge-2019/1.0.0/}. 
In brief, given $n$ patients' variables, $\{\mathcal{X}_{1}, \mathcal{X}_{2},\cdots, \mathcal{X}_{n}\}$, where the $i$-th patient's data is $\mathcal{X}_{i}=\{\boldsymbol{x}_{i,1}, \boldsymbol{x}_{i,2}, \cdots, \boldsymbol{x}_{i,m} \}$, $\boldsymbol{x}_{i,j}$ is the clinical features of $j$-th hour since patient $i$ entered ICU. 
For each 6-hour records, $\{\boldsymbol{x}_{i,k}, \boldsymbol{x}_{i,k+1}, \cdots, \boldsymbol{x}_{i,k+5}\}$, we aim to predict whether sepsis will occur for patient $i$ in the next 6 hours ,i.e., before $(k+11)$-th hour.

\vspace{+.5em}
\noindent \textbf{Cross-center Early Sepsis Recognition.}
When there are only limited EMRs per hospital, it is difficult to guarantee the model performance. 
In this paper, we focus on the collaborative learning of two participants (i.e., hospitals), to generate better health status representations for sepsis recognition. 
From the machine learning perspective, it can be viewed as a multi-task learning task \cite{zhang2021survey}.

\begin{figure*}
\begin{minipage}{.64\linewidth}
    \centering
    \includegraphics[width=0.95\linewidth]{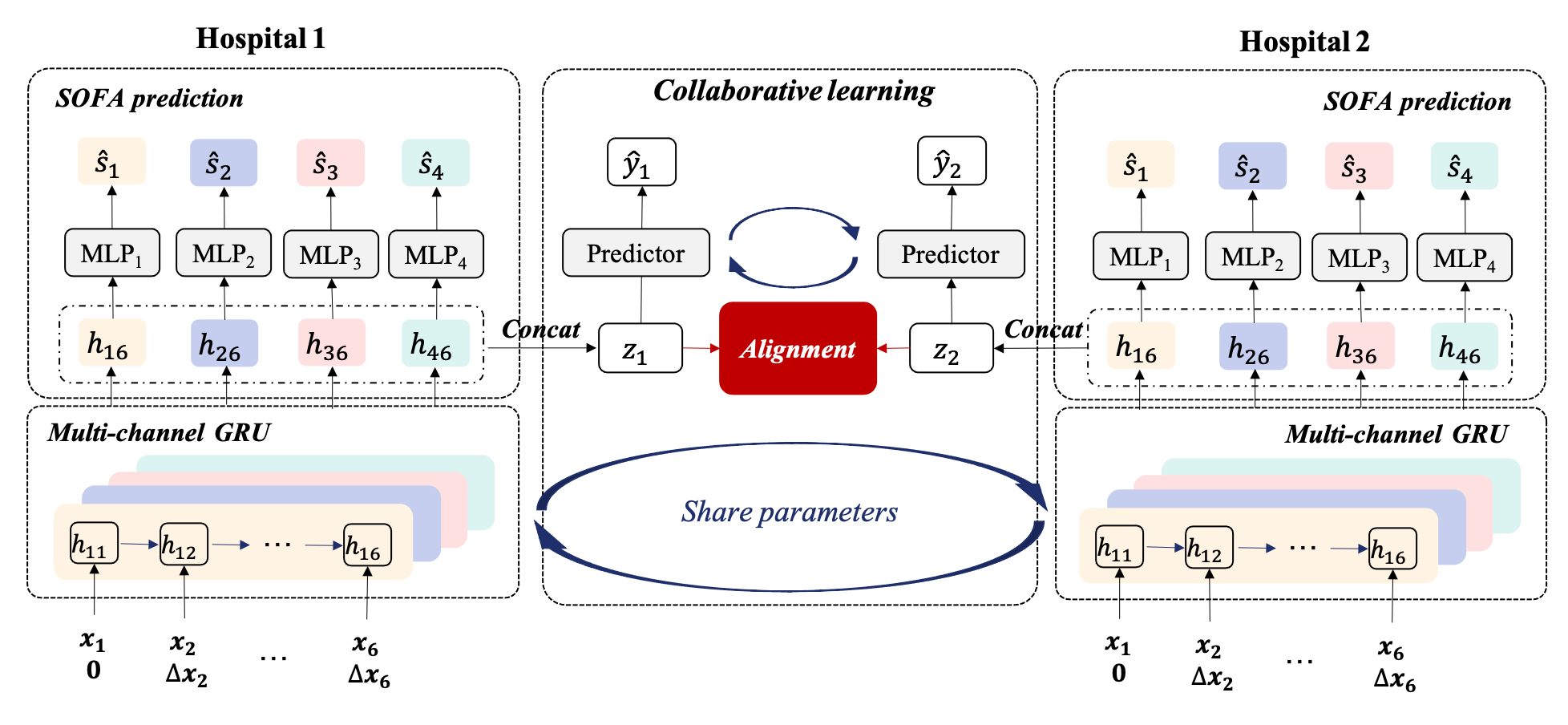}
    \vspace{-2em}
    \caption{The \method framework: (1) health status representation learning with SOFA prediction as the auxiliary task; (2) cross-center collaborative learning with model parameters sharing and feature distribution alignment in hidden space.}
    \label{fig:sofanet}
    \vspace{-1em}
\end{minipage}
\begin{minipage}{.32\linewidth}
  \centering
  \small
  \begin{tabular}[t]{lcc}
  \toprule
  &\textbf{MIMIC-III}          &\textbf{Challenge}\\
  \midrule
  \# patients         & 13379       & 4717    \\
  \# septic patients  & 2688         & 1441   \\
  Sepsis prevalence (\%)      &  20.09     &  30.55 \\
  \midrule
  \# records        & 69946         & 18616    \\
  \begin{tabular}[c]{@{}l@{}}\# records occur Sepsis \\ in next 6 hours \end{tabular} & 8751   & 2476  \\
  records with sepsis (\%) &  12.51  &  13.30 \\
  \midrule
  missing rate (\%) &  20.83   &   63.07  \\
  \bottomrule
  \end{tabular}
  \vspace{2em}
  \captionof{table}{The Statistics of Datasets}
  \label{table:statistic}
  \vspace{-1em}
\end{minipage}
\end{figure*}

\section{Methodology}

\subsection{Overview}

\method is proposed to achieve secure and compliant knowledge transfer when there is limited data in each hospital. 
Figure \ref{fig:sofanet} shows the overall framework, which contains two key parts, (i) \textit{Multi-channel GRU for health status representation learning} and (ii) \textit{Privacy-preserving Cross-center Collaborative Learning}.

\subsection{Health Status Representation Learning}
\label{sub:health_representation_learning}

Existing machine learning methods for early sepsis recognition usually input EMR variables and output a binary label on whether sepsis would occur \cite{fleuren2020machine,shashikumar2017early}.
For hospitals with limited patients' EMRs, this learning strategy may not perform well as the supervision signals are restricted to the small number of sepsis labels.

To increase the supervision signals for such hospitals without sufficient data, our proposed \textit{SofaNet} includes a novel set of auxiliary tasks by incorporating the medical expert knowledge from the sepsis diagnosis guideline \cite{10.1001/jama.2016.0287}.
Specifically, since SOFA >= 2 is one of the key conditions for sepsis diagnosis, SOFA score prediction has the potential to become an auxiliary task for early sepsis recognition.
Furthermore, SOFA scores can be calculated for six criteria including respiratory, cardiovascular, hepatic, renal, coagulation, and neurological systems. Hence, multiple new supervision signals on SOFA scores can be provided.

With this prior medical knowledge, we adopt a multi-channel GRU structure to embed the health status of each system individually.
Also, the dynamic changes of vital signs and laboratory variables can acutely reflect changes of a patient's status \cite{ma2020concare}. 
Therefore, given a 6-hour record of a patient, $\{\boldsymbol{x}_{1}, \boldsymbol{x}_{2}, \cdots, \boldsymbol{x}_{6}\}$, we compute the differential features with $\Delta \boldsymbol{x}_i = \boldsymbol{x}_i - \boldsymbol{x}_{i-1}$ when $i > 1$ and set $\Delta \boldsymbol{x}_{1} = \boldsymbol{0}$. 
As shown in Figure \ref{fig:sofanet}, we take the concatenation of original features and differential features as the input, i.e.,  $\{(\boldsymbol{x}_{1}, \boldsymbol{0}), (\boldsymbol{x}_{2}, \Delta \boldsymbol{x}_{2}),\cdots,$ $(\boldsymbol{x}_{6}, \Delta \boldsymbol{x}_{6})\}$. 
Due to the missing variables in the datasets,  SOFA scores can be precisely calculated for four systems (out of six), including coagulation, liver, cardiovascular, and renal \footnote{The features and SOFA scoring standard can refer to Table \ref{table:features} and Table \ref{table:sofa} respectively in Appendix.}.
Therefore, we build a 4-channel GRU feature extractor with the same input. 
For each channel, we take the last hidden state of GRU as the output, i.e., $\{h_{16}, h_{26}, h_{36}, h_{46}\}$.
As the SOFA score prediction is the auxiliary task, the loss function for each hospital can be written as
\begin{equation}
    \mathcal{L}_{local} = \mathcal{L}_{sepsis} + \alpha * \sum_{i=1}^{4} \mathcal{L}_{sofa_i}
\end{equation}
where $\mathcal{L}_{sepsis}$ and $\mathcal{L}_{sofa_i}$ are cross entropy, and $\alpha$ is set to $0.5$ \footnote{This hyperparameter is determined according to the experimental results, and the difference is small when using different values (from 0.5 to 1.0).}. 
In brief, in addition to the supervision signal of sepsis ($\mathcal{L}_{sepsis}$), we add four new supervision signals of SOFA scores ($\mathcal{L}_{sofa_i}, i=1...4$) to improve the learning robustness for hospitals with scarce data.

\subsection{Cross-center Collaborative Learning}
In addition to auxiliary tasks on SOFA prediction, we introduce a cross-center collaborative learning procedure for multiple hospitals.
Specifically, we expect that this collaborative procedure can enable data-scarce hospitals to benefit each other, so that they can be motivated to participate.
Note that as direct data sharing may violate data protection regulations such as GDPR,
we need to ensure that knowledge is shared while raw data is well protected.

Based on this idea, we design a cross-center collaborative learning mechanism, which achieves knowledge sharing by two ways: 
(i) \textit{model parameter sharing} in each iteration, which can be seen as the simplified version of the privacy-preserving federated learning algorithm,  \textit{FedAvg} \cite{mcmahan2017communication}, since there are only two participants and the central server is unnecessary;
(ii) \textit{feature distribution alignment} in hidden space, like domain adaptation \cite{pan2010domain}, to avoid diverged distributions between two hospitals, which may result in negative knowledge transfer.
Various alignment methods can be implemented, 
such as maximum mean discrepancy (MMD) \cite{tzeng2014deep} and adversarial training \cite{pmlr-v37-ganin15,peng2019federated}.
In our current implementation, we use the MMD method (denoted as $\textit{SofaNet}_{mmd}$) as it performs generally well in our experiments.
The loss function of $\textit{SofaNet}_{mmd}$ is 
\begin{equation}
    \mathcal{L} = \mathcal{L}_{local_1} + \mathcal{L}_{local_2} + \mathcal{L}_{mmd}
\end{equation}

\textbf{Privacy Protection}. 
In training $\textit{SofaNet}_{mmd}$, two hospitals only need to transfer intermediate results (i.e.,, model parameters and hidden representations) instead of raw data, which follows the privacy protection criteria in state-of-the-art federated domain adaptation methods \cite{peng2019federated}. 
Meanwhile, recent studies show that such intermediate results may also indirectly reveal raw data under certain conditions \cite{zhu2019deep}. In our future work, we plan to add advanced techniques, such as homomorphic encryption and differential privacy, to further protect the intermediate training results~\cite{lyu2020privacy}.

\section{Experiments}

\subsection{Experiment Setup}
We conduct our experiments on two widely-used real-life Sepsis recognition datasets, \textbf{MIMIC-III} \cite{johnson2016mimic} and the PhysioNet Computing in Cardiology Challenge 2019 (\textbf{Challenge}) \cite{reyna2019early}. 
As some patients' records have a large number of missing values, we screen out the patients whose missing value ratio is less than 80\%. 
For missing values, we fill in with data from the previous time point. If a feature at the initial time point is missing, we fill in with the mean value.
Through such preprocessing, we obtain the final data for experiments and keep the records of 10\% patients as test data. 
Some basic statistic information is enumerated in Table \ref{table:statistic}. 
We list the detailed features in Table \ref{table:features} (Appendix), including 3 demographic variables, 6 vital sign variables and 18 laboratory variables. The experiment environment is described in Appendix.

\subsection{Baselines}
In our experiments, we assume that there is only $x\%$ ($x=1$ by default) patients' records from MIMIC or Challenge for two hospitals, respectively, to simulate a data-scarce scenario.
To compare with our method \method, we implement two types of methods.
\begin{itemize}
    \item \textit{Local Learning}: each hospital trains a classifier only with its local medical records.
    \item \textit{Collaborative Learning}: two hospitals collaborate by techniques such as parameter sharing and finetuning.
\end{itemize}

For \textit{local learning}, we introduce several classical models widely used for sepsis recognition, including, \textit{Logistic Regression} (\textit{LR}) \cite{shashikumar2017early}, \textit{Neural Network} (\textit{NN}) \cite{futoma2017learning}, \textit{XGBoost}\cite{zabihi2019sepsis}, and \textit{GRU}\cite{chung2014empirical}. 
For \textit{collaborative learning}, we mainly compare with the state-of-the-art finetuning methods without raw data exchange \cite{ma2021distilling,kachuee2018ecg,leitner2021personalized}.
Following previous research using EMR data \cite{choi2018mime,ma2020concare,ma2021distilling}, model performance is assessed by the area under the receiver operating characteristic curve (AUROC), area under the precision-recall curve (AUPRC), and the minimum of precision and sensitivity (Min(Se,P+)).

\subsection{Experimental Results}

\subsubsection{Main Result.}
As shown in Table \ref{table:main_result}, \method outperforms in most metrics (except for the AUROC on MIMIC-III), demonstrating \method can learn a better representation through knowledge transfer while protecting the raw data. 
Concretely, compared to the best local methods, \method achieves a 0.9\% higher AUROC, a 5.39\% higher AUPRC, a 2.56\% higher min(Se,P+) on MIMIC-III dataset, and achieves a 7.70\% higher AUROC, a 50.22\% higher AUPRC and a 25.39\% higher min(Se, P+) on Challenge dataset.

\textbf{Effectiveness of Collaborative Learning}: By comparing the local methods and collaborative methods, we can conclude that learning knowledge from each other between different datasets can promote the prediction performance for their respective tasks. Also, we can observe that the improvement is more obvious on Challenge dataset, because of the relatively poorer data quality (i.e., smaller data size and higher missing rate shown in Table \ref{table:statistic}).

\textbf{Effectiveness of Knowledge-guide Multi-channel GRU}: Compared to \textit{GRU} and $\textit{SofaNet}_{lc}$ \textit{w.o. MC} model, $\textit{SofaNet}_{lc}$ achieves a better performance during local training on two datasets. This indicates that taking SOFA values prediction as the auxiliary task is helpful for data-scarce hospitals' early sepsis recognition.

\textbf{Future direction on combining \textit{XGBoost} and \textit{SofaNet}}: 
Among local methods, \textit{XGBoost} performs the best, outperforming all the deep learning methods including $\textit{SofaNet}_{lc}$.
This is actually consistent with the competition results of the PhysioNet Challenge, where the top three teams all use \textit{XGBoost}-like ensemble methods \cite{morrill2019signature,du2019automated,zabihi2019sepsis}.
A promising direction would be how to combine the local learning capacity of \textit{XGBoost} and the collaboration power of $\textit{SofaNet}_{mmd}$.
We have attempted to use \textit{XGBoost} on the collaboratively aligned representations of $\textit{SofaNet}_{mmd}$ (i.e., $z_1$ and $z_2$ in Figure~\ref{fig:sofanet}) and observed certain improvements (although not stable).
This confirms the feasibility of combining these two methods, and we highly believe that more advanced combination techniques can be developed in the future for significant improvements.

\begin{table}[t]
\footnotesize
\caption{Early Sepsis Recognition Performance with Only 1\% Patients as the Training Set}
\vspace{-1em}
\label{table:main_result}
\begin{threeparttable}
\resizebox{\columnwidth}{!}{%
\begin{tabular}{lcccccc}
\toprule
 &
  \multicolumn{3}{c}{\textbf{MIMIC-III}} &
  \multicolumn{3}{c}{\textbf{Challenge}} \\ 
  \cmidrule(r){2-4}  \cmidrule(r){5-7}
\textit{} &
  \multicolumn{1}{c}{AUROC} &
  \multicolumn{1}{c}{AUPRC} &
  \multicolumn{1}{c}{Min(Se, P+)} &
  \multicolumn{1}{c}{AUROC} &
  \multicolumn{1}{c}{AUPRC} &
  \multicolumn{1}{c}{Min(Se, P+)} \\ \midrule
\multicolumn{7}{l}{\textit{\textbf{Local Learning}}}                 \\
\multirow{2}{*}{\textit{LR} }  &  0.8987   &  0.6583  &  0.6022  &  0.5724   &  0.1715 &  0.2279  \\
                               &  (0.002)  &  (0.003) &  (0.001) &  (0.002)  &  (0.001) & (0.002) \\
\multirow{2}{*}{\textit{NN}}  &  0.8625   &  0.5986 &  0.5452 &  0.4745   &  0.1334 &  0.1457  \\
                              & (0.008)  & (0.014) & (0.006) & (0.009)  &  (0.004)  &  (0.007) \\
\multirow{2}{*}{\textit{XGBoost}} &  \underline{0.9107}   &  \underline{0.6829}  &  \underline{0.6285}  &  \underline{0.7588}   &  \underline{0.2676}  &  \underline{0.3292} \\
                                  &  (0.003) &  (0.006) &  (0.009) &  (0.009) &  (0.007) &  (0.008)  \\
\multirow{2}{*}{\textit{GRU}} &  0.8992  &  0.6639  &  0.5994 &  0.6283 &  0.2247 &  0.2667 \\
                              &  (0.006) &  (0.012) &  (0.008) & (0.044)  & (0.019) & (0.025) \\
\multirow{2}{*}{$\textit{SofaNet}_{lc}$ \textit{w.o. MC}}  &  0.9046 & 0.6726 &  0.6004  &  0.6647 &  0.2476  &  0.2941  \\ 
                                                              & (0.006) & (0.104) & (0.009) &  (0.052) & (0.043) & (0.063) \\
\multirow{2}{*}{$\textit{SofaNet}_{lc}$}   &  0.9067  &  0.6807  &  0.6171  &  0.6895  &  0.2609  &  0.2982  \\
                                              & (0.006) & (0.017)  & (0.009)  &  (0.036)  & (0.019)  & (0.015) \\
\midrule
\multicolumn{7}{l}{\textit{\textbf{Collaborative Learning}}} \\
\multirow{2}{*}{$\textit{Finetune}$}  &  \textbf{0.9216}   &  0.7113  &  0.6314  &  0.7955  &  0.3161  &  0.3831  \\
                                      &  (0.006) & (0.028) & (0.026) & (0.010)  & (0.016)  & (0.020)  \\
\multirow{2}{*}{$\textit{SofaNet}_{mmd}$}  &  0.9187  &  \textbf{0.7197} &  \textbf{0.6446}  & \textbf{0.8172}  & \textbf{0.4020}  &  \textbf{0.4128}  \\ 
                                           & (0.006) & (0.016) & (0.008) & (0.007) & (0.023) & (0.025) \\
\bottomrule
\end{tabular}%
}
\end{threeparttable}
\begin{tablenotes}
    \footnotesize
    \item 1. Values in ``( )'' denote the standard deviation of five experiments' results;
    \item 2. \textbf{Bold} denotes the best-performed ones of the task; 
    \item 3. Underline denotes the best-performed ones in local learning;
    \item 4. $\textit{SofaNet}_{lc}$ means \textit{SofaNet} without collaborative training; 
    \item 5. $\textit{SofaNet}_{lc}$ \textit{ w.o. MC} means $\textit{SofaNet}_{lc}$ without multi-channel GRU.
\end{tablenotes}
\vspace{-1.5em}
\end{table}

\subsubsection{Varying the Size of Training Set.}
\begin{figure}[t]
    \centering
    \subfigure[MIMIC-III]{
    \includegraphics[width=0.47\linewidth]{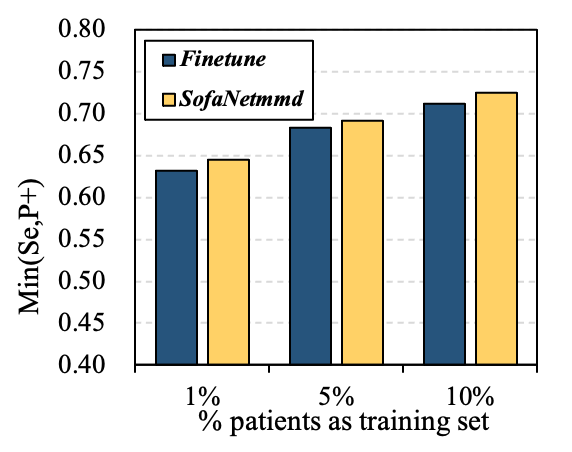}
    }
    \subfigure[Challenge]{
    \includegraphics[width=0.47\linewidth]{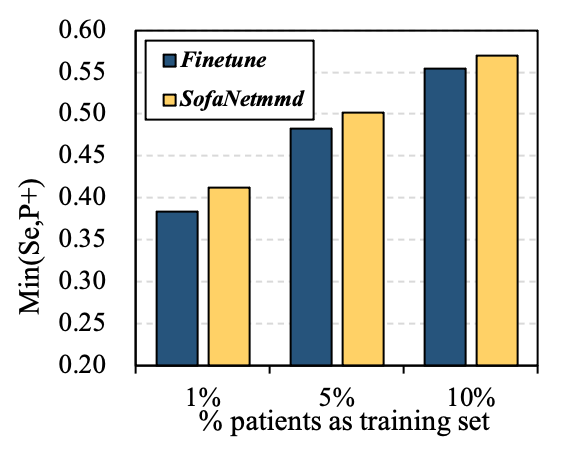}
    }
    \vspace{-1em}
    \caption{Performance under different training data volume.}
    \label{fig:train_data_size}
    \vspace{-2em}
\end{figure}

Fixing the test data, we adjust the data size of training set, i.e., sample 1\%, 5\%, 10\% patients and use their medical records as the training data.
Figure \ref{fig:train_data_size} shows the Min(Se, P+) values of MIMIC-III and Challenge test data under different training data sizes respectively.
The Min(Se, P+) values rise as the training data size increases, and $\method_{mmd}$ consistently outperforms \textit{Finetune}.
The performance gap between \method and \textit{Finetune} is more considerable in smaller datasets, which indicates the capability of our proposed collaborative learning mechanism to alleviate the data insufficiency problem.

\section{Conclusion}

In this paper, we propose a privacy-preserving cross-center collaborative learning method, \method, for early sepsis recognition. The experimental results on two datasets show the effectiveness of \method, especially for hospitals with scarce data. 
We mainly achieve raw data protection through parameters sharing and health status representation alignment. 
Also, we find that \textit{XGBoost} performs well in local training. 
It is worth thinking about how to further integrate \textit{XGBoost} into \method to get better results.

\textbf{Limitation \& Future Work}.
We focus on early sepsis recognition by two-hospital collaborative learning. 
However, this method can be easily extended to $n$-hospital scenario: first, every two hospitals can do collaborative learning for a better health status representation; second, one hospital can combine (e.g., concatenation) all the collaboratively learned representations with other hospitals for its own patients’ early sepsis recognition.
For the representative disease, sepsis, we used the Sepsis-3 guideline \cite{10.1001/jama.2016.0287} to guide our method design. There are also diagnostic guidelines for different diseases, like KDIGO for acute kidney injury \cite{disease2012improving}.
With the corresponding knowledge, we can design models like \method.
This idea to inject domain knowledge can be generalized to other domains.

\begin{acks}
We appreciate Yanlin Chen for her discussion of the model design idea.
This research was supported by National Key R\&D Program of China (2020AAA0109401) and NSFC Grants no. 72071125, 61972008, and 72031001.
\end{acks}

\bibliographystyle{ACM-Reference-Format}
\bibliography{ref}


\begin{thebibliography}{42}


\ifx \showCODEN    \undefined \def \showCODEN     #1{\unskip}     \fi
\ifx \showDOI      \undefined \def \showDOI       #1{#1}\fi
\ifx \showISBNx    \undefined \def \showISBNx     #1{\unskip}     \fi
\ifx \showISBNxiii \undefined \def \showISBNxiii  #1{\unskip}     \fi
\ifx \showISSN     \undefined \def \showISSN      #1{\unskip}     \fi
\ifx \showLCCN     \undefined \def \showLCCN      #1{\unskip}     \fi
\ifx \shownote     \undefined \def \shownote      #1{#1}          \fi
\ifx \showarticletitle \undefined \def \showarticletitle #1{#1}   \fi
\ifx \showURL      \undefined \def \showURL       {\relax}        \fi
\providecommand\bibfield[2]{#2}
\providecommand\bibinfo[2]{#2}
\providecommand\natexlab[1]{#1}
\providecommand\showeprint[2][]{arXiv:#2}

\bibitem[Choi et~al\mbox{.}(2018)]%
        {choi2018mime}
\bibfield{author}{\bibinfo{person}{Edward Choi}, \bibinfo{person}{Cao Xiao},
  \bibinfo{person}{Walter Stewart}, {and} \bibinfo{person}{Jimeng Sun}.}
  \bibinfo{year}{2018}\natexlab{}.
\newblock \showarticletitle{Mime: Multilevel medical embedding of electronic
  health records for predictive healthcare}.
\newblock \bibinfo{journal}{\emph{Advances in neural information processing
  systems}}  \bibinfo{volume}{31} (\bibinfo{year}{2018}).
\newblock


\bibitem[Chung et~al\mbox{.}(2014)]%
        {chung2014empirical}
\bibfield{author}{\bibinfo{person}{Junyoung Chung}, \bibinfo{person}{Caglar
  Gulcehre}, \bibinfo{person}{KyungHyun Cho}, {and} \bibinfo{person}{Yoshua
  Bengio}.} \bibinfo{year}{2014}\natexlab{}.
\newblock \showarticletitle{Empirical evaluation of gated recurrent neural
  networks on sequence modeling}.
\newblock \bibinfo{journal}{\emph{arXiv preprint arXiv:1412.3555}}
  (\bibinfo{year}{2014}).
\newblock


\bibitem[Disease(2012)]%
        {disease2012improving}
\bibfield{author}{\bibinfo{person}{KJKIS Disease}.}
  \bibinfo{year}{2012}\natexlab{}.
\newblock \showarticletitle{Improving global outcomes (KDIGO) acute kidney
  injury work group: KDIGO clinical practice guideline for acute kidney
  injury}.
\newblock \bibinfo{journal}{\emph{Kidney Int Suppl}} \bibinfo{volume}{2},
  \bibinfo{number}{1} (\bibinfo{year}{2012}), \bibinfo{pages}{1--138}.
\newblock


\bibitem[Du et~al\mbox{.}(2019)]%
        {du2019automated}
\bibfield{author}{\bibinfo{person}{John~Anda Du}, \bibinfo{person}{Nadi Sadr},
  {and} \bibinfo{person}{Philip de Chazal}.} \bibinfo{year}{2019}\natexlab{}.
\newblock \showarticletitle{Automated prediction of sepsis onset using gradient
  boosted decision trees}. In \bibinfo{booktitle}{\emph{2019 Computing in
  Cardiology (CinC)}}. IEEE, \bibinfo{pages}{Page--1}.
\newblock


\bibitem[Ferrer et~al\mbox{.}(2008)]%
        {ferrer2008improvement}
\bibfield{author}{\bibinfo{person}{Ricard Ferrer}, \bibinfo{person}{Antonio
  Artigas}, \bibinfo{person}{Mitchell~M Levy}, \bibinfo{person}{Jesus Blanco},
  \bibinfo{person}{Gumersindo Gonzalez-Diaz}, \bibinfo{person}{Jos{\'e}
  Garnacho-Montero}, \bibinfo{person}{Jordi Ib{\'a}{\~n}ez},
  \bibinfo{person}{Eduardo Palencia}, \bibinfo{person}{Manuel Quintana},
  \bibinfo{person}{Mar{\'\i}a~Victoria De~La Torre-Prados}, {et~al\mbox{.}}}
  \bibinfo{year}{2008}\natexlab{}.
\newblock \showarticletitle{Improvement in process of care and outcome after a
  multicenter severe sepsis educational program in Spain}.
\newblock \bibinfo{journal}{\emph{Jama}} \bibinfo{volume}{299},
  \bibinfo{number}{19} (\bibinfo{year}{2008}), \bibinfo{pages}{2294--2303}.
\newblock


\bibitem[Fleuren et~al\mbox{.}(2020)]%
        {fleuren2020machine}
\bibfield{author}{\bibinfo{person}{Lucas~M Fleuren}, \bibinfo{person}{Thomas~LT
  Klausch}, \bibinfo{person}{Charlotte~L Zwager}, \bibinfo{person}{Linda~J
  Schoonmade}, \bibinfo{person}{Tingjie Guo}, \bibinfo{person}{Luca~F
  Roggeveen}, \bibinfo{person}{Eleonora~L Swart}, \bibinfo{person}{Armand~RJ
  Girbes}, \bibinfo{person}{Patrick Thoral}, \bibinfo{person}{Ari Ercole},
  {et~al\mbox{.}}} \bibinfo{year}{2020}\natexlab{}.
\newblock \showarticletitle{Machine learning for the prediction of sepsis: a
  systematic review and meta-analysis of diagnostic test accuracy}.
\newblock \bibinfo{journal}{\emph{Intensive care medicine}}
  \bibinfo{volume}{46}, \bibinfo{number}{3} (\bibinfo{year}{2020}),
  \bibinfo{pages}{383--400}.
\newblock


\bibitem[Futoma et~al\mbox{.}(2017)]%
        {futoma2017learning}
\bibfield{author}{\bibinfo{person}{Joseph Futoma}, \bibinfo{person}{Sanjay
  Hariharan}, {and} \bibinfo{person}{Katherine Heller}.}
  \bibinfo{year}{2017}\natexlab{}.
\newblock \showarticletitle{Learning to detect sepsis with a multitask Gaussian
  process RNN classifier}. In \bibinfo{booktitle}{\emph{International
  Conference on Machine Learning}}. PMLR, \bibinfo{pages}{1174--1182}.
\newblock


\bibitem[Ganin and Lempitsky(2015)]%
        {pmlr-v37-ganin15}
\bibfield{author}{\bibinfo{person}{Yaroslav Ganin} {and}
  \bibinfo{person}{Victor Lempitsky}.} \bibinfo{year}{2015}\natexlab{}.
\newblock \showarticletitle{Unsupervised Domain Adaptation by Backpropagation}.
  In \bibinfo{booktitle}{\emph{Proceedings of the 32nd International Conference
  on Machine Learning}} \emph{(\bibinfo{series}{Proceedings of Machine Learning
  Research}, Vol.~\bibinfo{volume}{37})},
  \bibfield{editor}{\bibinfo{person}{Francis Bach} {and} \bibinfo{person}{David
  Blei}} (Eds.). \bibinfo{publisher}{PMLR}, \bibinfo{address}{Lille, France},
  \bibinfo{pages}{1180--1189}.
\newblock


\bibitem[Goldberger et~al\mbox{.}(2000)]%
        {goldberger2000physiobank}
\bibfield{author}{\bibinfo{person}{Ary~L Goldberger}, \bibinfo{person}{Luis~AN
  Amaral}, \bibinfo{person}{Leon Glass}, \bibinfo{person}{Jeffrey~M Hausdorff},
  \bibinfo{person}{Plamen~Ch Ivanov}, \bibinfo{person}{Roger~G Mark},
  \bibinfo{person}{Joseph~E Mietus}, \bibinfo{person}{George~B Moody},
  \bibinfo{person}{Chung-Kang Peng}, {and} \bibinfo{person}{H~Eugene Stanley}.}
  \bibinfo{year}{2000}\natexlab{}.
\newblock \showarticletitle{PhysioBank, PhysioToolkit, and PhysioNet:
  components of a new research resource for complex physiologic signals}.
\newblock \bibinfo{journal}{\emph{circulation}} \bibinfo{volume}{101},
  \bibinfo{number}{23} (\bibinfo{year}{2000}), \bibinfo{pages}{e215--e220}.
\newblock


\bibitem[Gupta et~al\mbox{.}(2020)]%
        {gupta2020transfer}
\bibfield{author}{\bibinfo{person}{Priyanka Gupta}, \bibinfo{person}{Pankaj
  Malhotra}, \bibinfo{person}{Jyoti Narwariya}, \bibinfo{person}{Lovekesh Vig},
  {and} \bibinfo{person}{Gautam Shroff}.} \bibinfo{year}{2020}\natexlab{}.
\newblock \showarticletitle{Transfer learning for clinical time series analysis
  using deep neural networks}.
\newblock \bibinfo{journal}{\emph{Journal of Healthcare Informatics Research}}
  \bibinfo{volume}{4}, \bibinfo{number}{2} (\bibinfo{year}{2020}),
  \bibinfo{pages}{112--137}.
\newblock


\bibitem[Johnson et~al\mbox{.}(2016)]%
        {johnson2016mimic}
\bibfield{author}{\bibinfo{person}{Alistair~EW Johnson}, \bibinfo{person}{Tom~J
  Pollard}, \bibinfo{person}{Lu Shen}, \bibinfo{person}{H~Lehman Li-Wei},
  \bibinfo{person}{Mengling Feng}, \bibinfo{person}{Mohammad Ghassemi},
  \bibinfo{person}{Benjamin Moody}, \bibinfo{person}{Peter Szolovits},
  \bibinfo{person}{Leo~Anthony Celi}, {and} \bibinfo{person}{Roger~G Mark}.}
  \bibinfo{year}{2016}\natexlab{}.
\newblock \showarticletitle{MIMIC-III, a freely accessible critical care
  database}.
\newblock \bibinfo{journal}{\emph{Scientific data}} \bibinfo{volume}{3},
  \bibinfo{number}{1} (\bibinfo{year}{2016}), \bibinfo{pages}{1--9}.
\newblock


\bibitem[Ju et~al\mbox{.}(2020)]%
        {ju2020privacy}
\bibfield{author}{\bibinfo{person}{Ce Ju}, \bibinfo{person}{Ruihui Zhao},
  \bibinfo{person}{Jichao Sun}, \bibinfo{person}{Xiguang Wei},
  \bibinfo{person}{Bo Zhao}, \bibinfo{person}{Yang Liu},
  \bibinfo{person}{Hongshan Li}, \bibinfo{person}{Tianjian Chen},
  \bibinfo{person}{Xinwei Zhang}, \bibinfo{person}{Dashan Gao},
  {et~al\mbox{.}}} \bibinfo{year}{2020}\natexlab{}.
\newblock \showarticletitle{Privacy-preserving technology to help millions of
  people: Federated prediction model for stroke prevention}.
\newblock \bibinfo{journal}{\emph{arXiv preprint arXiv:2006.10517}}
  (\bibinfo{year}{2020}).
\newblock


\bibitem[Kachuee et~al\mbox{.}(2018)]%
        {kachuee2018ecg}
\bibfield{author}{\bibinfo{person}{Mohammad Kachuee}, \bibinfo{person}{Shayan
  Fazeli}, {and} \bibinfo{person}{Majid Sarrafzadeh}.}
  \bibinfo{year}{2018}\natexlab{}.
\newblock \showarticletitle{Ecg heartbeat classification: A deep transferable
  representation}. In \bibinfo{booktitle}{\emph{2018 IEEE international
  conference on healthcare informatics (ICHI)}}. IEEE,
  \bibinfo{pages}{443--444}.
\newblock


\bibitem[Karpatne et~al\mbox{.}(2018)]%
        {karpatne2018machine}
\bibfield{author}{\bibinfo{person}{Anuj Karpatne}, \bibinfo{person}{Imme
  Ebert-Uphoff}, \bibinfo{person}{Sai Ravela}, \bibinfo{person}{Hassan~Ali
  Babaie}, {and} \bibinfo{person}{Vipin Kumar}.}
  \bibinfo{year}{2018}\natexlab{}.
\newblock \showarticletitle{Machine learning for the geosciences: Challenges
  and opportunities}.
\newblock \bibinfo{journal}{\emph{IEEE Transactions on Knowledge and Data
  Engineering}} \bibinfo{volume}{31}, \bibinfo{number}{8}
  (\bibinfo{year}{2018}), \bibinfo{pages}{1544--1554}.
\newblock


\bibitem[Kingma and Ba(2014)]%
        {kingma2014adam}
\bibfield{author}{\bibinfo{person}{Diederik~P Kingma} {and}
  \bibinfo{person}{Jimmy Ba}.} \bibinfo{year}{2014}\natexlab{}.
\newblock \showarticletitle{Adam: A method for stochastic optimization}.
\newblock \bibinfo{journal}{\emph{arXiv preprint arXiv:1412.6980}}
  (\bibinfo{year}{2014}).
\newblock


\bibitem[Kumar et~al\mbox{.}(2006)]%
        {kumar2006duration}
\bibfield{author}{\bibinfo{person}{Anand Kumar}, \bibinfo{person}{Daniel
  Roberts}, \bibinfo{person}{Kenneth~E Wood}, \bibinfo{person}{Bruce Light},
  \bibinfo{person}{Joseph~E Parrillo}, \bibinfo{person}{Satendra Sharma},
  \bibinfo{person}{Robert Suppes}, \bibinfo{person}{Daniel Feinstein},
  \bibinfo{person}{Sergio Zanotti}, \bibinfo{person}{Leo Taiberg},
  {et~al\mbox{.}}} \bibinfo{year}{2006}\natexlab{}.
\newblock \showarticletitle{Duration of hypotension before initiation of
  effective antimicrobial therapy is the critical determinant of survival in
  human septic shock}.
\newblock \bibinfo{journal}{\emph{Critical care medicine}}
  \bibinfo{volume}{34}, \bibinfo{number}{6} (\bibinfo{year}{2006}),
  \bibinfo{pages}{1589--1596}.
\newblock


\bibitem[Lee et~al\mbox{.}(2012)]%
        {lee2012adapting}
\bibfield{author}{\bibinfo{person}{Gyemin Lee}, \bibinfo{person}{Ilan
  Rubinfeld}, {and} \bibinfo{person}{Zeeshan Syed}.}
  \bibinfo{year}{2012}\natexlab{}.
\newblock \showarticletitle{Adapting surgical models to individual hospitals
  using transfer learning}. In \bibinfo{booktitle}{\emph{2012 IEEE 12th
  international conference on data mining workshops}}. IEEE,
  \bibinfo{pages}{57--63}.
\newblock


\bibitem[Leitner et~al\mbox{.}(2021)]%
        {leitner2021personalized}
\bibfield{author}{\bibinfo{person}{Jared Leitner}, \bibinfo{person}{Po-Han
  Chiang}, {and} \bibinfo{person}{Sujit Dey}.} \bibinfo{year}{2021}\natexlab{}.
\newblock \showarticletitle{Personalized blood pressure estimation using
  photoplethysmography: A transfer learning approach}.
\newblock \bibinfo{journal}{\emph{IEEE Journal of Biomedical and Health
  Informatics}} \bibinfo{volume}{26}, \bibinfo{number}{1}
  (\bibinfo{year}{2021}), \bibinfo{pages}{218--228}.
\newblock


\bibitem[Levy et~al\mbox{.}(2003)]%
        {levy_2001_2003}
\bibfield{author}{\bibinfo{person}{Mitchell~M. Levy},
  \bibinfo{person}{Mitchell~P. Fink}, \bibinfo{person}{John~C. Marshall},
  \bibinfo{person}{Edward Abraham}, \bibinfo{person}{Derek Angus},
  \bibinfo{person}{Deborah Cook}, \bibinfo{person}{Jonathan Cohen},
  \bibinfo{person}{Steven~M. Opal}, \bibinfo{person}{Jean-Louis Vincent},
  \bibinfo{person}{Graham Ramsay}, {and} \bibinfo{person}{{for the
  International Sepsis Definitions Conference}}.}
  \bibinfo{year}{2003}\natexlab{}.
\newblock \showarticletitle{2001 {SCCM}/{ESICM}/{ACCP}/{ATS}/{SIS}
  International Sepsis Definitions Conference}.
\newblock  \bibinfo{volume}{29}, \bibinfo{number}{4} (\bibinfo{year}{2003}),
  \bibinfo{pages}{530--538}.
\newblock
\showISSN{1432-1238}


\bibitem[Li et~al\mbox{.}(2020)]%
        {li2020time}
\bibfield{author}{\bibinfo{person}{Xiang Li}, \bibinfo{person}{Xiao Xu},
  \bibinfo{person}{Fei Xie}, \bibinfo{person}{Xian Xu}, \bibinfo{person}{Yuyao
  Sun}, \bibinfo{person}{Xiaoshuang Liu}, \bibinfo{person}{Xiaoyu Jia},
  \bibinfo{person}{Yanni Kang}, \bibinfo{person}{Lixin Xie},
  \bibinfo{person}{Fei Wang}, {et~al\mbox{.}}} \bibinfo{year}{2020}\natexlab{}.
\newblock \showarticletitle{A Time-Phased Machine Learning Model for Real-Time
  Prediction of Sepsis in Critical Care}.
\newblock \bibinfo{journal}{\emph{Critical Care Medicine}}
  \bibinfo{volume}{48}, \bibinfo{number}{10} (\bibinfo{year}{2020}),
  \bibinfo{pages}{e884--e888}.
\newblock


\bibitem[Lyu et~al\mbox{.}(2020)]%
        {lyu2020privacy}
\bibfield{author}{\bibinfo{person}{Lingjuan Lyu}, \bibinfo{person}{Han Yu},
  \bibinfo{person}{Xingjun Ma}, \bibinfo{person}{Lichao Sun},
  \bibinfo{person}{Jun Zhao}, \bibinfo{person}{Qiang Yang}, {and}
  \bibinfo{person}{Philip~S Yu}.} \bibinfo{year}{2020}\natexlab{}.
\newblock \showarticletitle{Privacy and robustness in federated learning:
  Attacks and defenses}.
\newblock \bibinfo{journal}{\emph{arXiv preprint arXiv:2012.06337}}
  (\bibinfo{year}{2020}).
\newblock


\bibitem[Ma et~al\mbox{.}(2021)]%
        {ma2021distilling}
\bibfield{author}{\bibinfo{person}{Liantao Ma}, \bibinfo{person}{Xinyu Ma},
  \bibinfo{person}{Junyi Gao}, \bibinfo{person}{Xianfeng Jiao},
  \bibinfo{person}{Zhihao Yu}, \bibinfo{person}{Chaohe Zhang},
  \bibinfo{person}{Wenjie Ruan}, \bibinfo{person}{Yasha Wang},
  \bibinfo{person}{Wen Tang}, {and} \bibinfo{person}{Jiangtao Wang}.}
  \bibinfo{year}{2021}\natexlab{}.
\newblock \showarticletitle{Distilling knowledge from publicly available online
  EMR data to emerging epidemic for prognosis}. In
  \bibinfo{booktitle}{\emph{Proceedings of the Web Conference 2021}}.
  \bibinfo{pages}{3558--3568}.
\newblock


\bibitem[Ma et~al\mbox{.}(2020)]%
        {ma2020concare}
\bibfield{author}{\bibinfo{person}{Liantao Ma}, \bibinfo{person}{Chaohe Zhang},
  \bibinfo{person}{Yasha Wang}, \bibinfo{person}{Wenjie Ruan},
  \bibinfo{person}{Jiangtao Wang}, \bibinfo{person}{Wen Tang},
  \bibinfo{person}{Xinyu Ma}, \bibinfo{person}{Xin Gao}, {and}
  \bibinfo{person}{Junyi Gao}.} \bibinfo{year}{2020}\natexlab{}.
\newblock \showarticletitle{Concare: Personalized clinical feature embedding
  via capturing the healthcare context}. In
  \bibinfo{booktitle}{\emph{Proceedings of the AAAI Conference on Artificial
  Intelligence}}, Vol.~\bibinfo{volume}{34}. \bibinfo{pages}{833--840}.
\newblock


\bibitem[McMahan et~al\mbox{.}(2017)]%
        {mcmahan2017communication}
\bibfield{author}{\bibinfo{person}{Brendan McMahan}, \bibinfo{person}{Eider
  Moore}, \bibinfo{person}{Daniel Ramage}, \bibinfo{person}{Seth Hampson},
  {and} \bibinfo{person}{Blaise~Aguera y Arcas}.}
  \bibinfo{year}{2017}\natexlab{}.
\newblock \showarticletitle{Communication-efficient learning of deep networks
  from decentralized data}. In \bibinfo{booktitle}{\emph{Artificial
  intelligence and statistics}}. PMLR, \bibinfo{pages}{1273--1282}.
\newblock


\bibitem[Miotto et~al\mbox{.}(2018)]%
        {miotto2018deep}
\bibfield{author}{\bibinfo{person}{Riccardo Miotto}, \bibinfo{person}{Fei
  Wang}, \bibinfo{person}{Shuang Wang}, \bibinfo{person}{Xiaoqian Jiang}, {and}
  \bibinfo{person}{Joel~T Dudley}.} \bibinfo{year}{2018}\natexlab{}.
\newblock \showarticletitle{Deep learning for healthcare: review, opportunities
  and challenges}.
\newblock \bibinfo{journal}{\emph{Briefings in bioinformatics}}
  \bibinfo{volume}{19}, \bibinfo{number}{6} (\bibinfo{year}{2018}),
  \bibinfo{pages}{1236--1246}.
\newblock


\bibitem[Morrill et~al\mbox{.}(2019)]%
        {morrill2019signature}
\bibfield{author}{\bibinfo{person}{James Morrill}, \bibinfo{person}{Andrey
  Kormilitzin}, \bibinfo{person}{Alejo Nevado-Holgado},
  \bibinfo{person}{Sumanth Swaminathan}, \bibinfo{person}{Sam Howison}, {and}
  \bibinfo{person}{Terry Lyons}.} \bibinfo{year}{2019}\natexlab{}.
\newblock \showarticletitle{The signature-based model for early detection of
  sepsis from electronic health records in the intensive care unit}. In
  \bibinfo{booktitle}{\emph{2019 Computing in Cardiology (CinC)}}. IEEE,
  \bibinfo{pages}{Page--1}.
\newblock


\bibitem[of~Chest~Physicians et~al\mbox{.}(1992)]%
        {american1992american}
\bibfield{author}{\bibinfo{person}{American~College of Chest~Physicians},
  \bibinfo{person}{Society of Critical Care Medicine Consensus
  Conference~Committee}, {et~al\mbox{.}}} \bibinfo{year}{1992}\natexlab{}.
\newblock \showarticletitle{American College of Chest Physicians/Society of
  Critical Care Medicine Consensus Conference: definitions for sepsis and organ
  failure and guidelines for the use of innovative therapies in sepsis}.
\newblock \bibinfo{journal}{\emph{Crit. Care Med}}  \bibinfo{volume}{20}
  (\bibinfo{year}{1992}), \bibinfo{pages}{864--874}.
\newblock


\bibitem[Pan et~al\mbox{.}(2010)]%
        {pan2010domain}
\bibfield{author}{\bibinfo{person}{Sinno~Jialin Pan}, \bibinfo{person}{Ivor~W
  Tsang}, \bibinfo{person}{James~T Kwok}, {and} \bibinfo{person}{Qiang Yang}.}
  \bibinfo{year}{2010}\natexlab{}.
\newblock \showarticletitle{Domain adaptation via transfer component analysis}.
\newblock \bibinfo{journal}{\emph{IEEE Transactions on Neural Networks}}
  \bibinfo{volume}{22}, \bibinfo{number}{2} (\bibinfo{year}{2010}),
  \bibinfo{pages}{199--210}.
\newblock


\bibitem[Pan and Yang(2009)]%
        {pan2009survey}
\bibfield{author}{\bibinfo{person}{Sinno~Jialin Pan} {and}
  \bibinfo{person}{Qiang Yang}.} \bibinfo{year}{2009}\natexlab{}.
\newblock \showarticletitle{A survey on transfer learning}.
\newblock \bibinfo{journal}{\emph{IEEE Transactions on knowledge and data
  engineering}} \bibinfo{volume}{22}, \bibinfo{number}{10}
  (\bibinfo{year}{2009}), \bibinfo{pages}{1345--1359}.
\newblock


\bibitem[Peng et~al\mbox{.}(2019)]%
        {peng2019federated}
\bibfield{author}{\bibinfo{person}{Xingchao Peng}, \bibinfo{person}{Zijun
  Huang}, \bibinfo{person}{Yizhe Zhu}, {and} \bibinfo{person}{Kate Saenko}.}
  \bibinfo{year}{2019}\natexlab{}.
\newblock \showarticletitle{Federated Adversarial Domain Adaptation}. In
  \bibinfo{booktitle}{\emph{International Conference on Learning
  Representations}}.
\newblock


\bibitem[Price and Cohen(2019)]%
        {price2019privacy}
\bibfield{author}{\bibinfo{person}{W~Nicholson Price} {and}
  \bibinfo{person}{I~Glenn Cohen}.} \bibinfo{year}{2019}\natexlab{}.
\newblock \showarticletitle{Privacy in the age of medical big data}.
\newblock \bibinfo{journal}{\emph{Nature medicine}} \bibinfo{volume}{25},
  \bibinfo{number}{1} (\bibinfo{year}{2019}), \bibinfo{pages}{37--43}.
\newblock


\bibitem[Reyna et~al\mbox{.}(2019)]%
        {reyna2019early}
\bibfield{author}{\bibinfo{person}{Matthew~A Reyna}, \bibinfo{person}{Chris
  Josef}, \bibinfo{person}{Salman Seyedi}, \bibinfo{person}{Russell Jeter},
  \bibinfo{person}{Supreeth~P Shashikumar}, \bibinfo{person}{M~Brandon
  Westover}, \bibinfo{person}{Ashish Sharma}, \bibinfo{person}{Shamim Nemati},
  {and} \bibinfo{person}{Gari~D Clifford}.} \bibinfo{year}{2019}\natexlab{}.
\newblock \showarticletitle{Early prediction of sepsis from clinical data: the
  PhysioNet/Computing in Cardiology Challenge 2019}. In
  \bibinfo{booktitle}{\emph{2019 Computing in Cardiology (CinC)}}. IEEE,
  \bibinfo{pages}{Page--1}.
\newblock


\bibitem[Rudd et~al\mbox{.}(2020)]%
        {rudd2020global}
\bibfield{author}{\bibinfo{person}{Kristina~E Rudd},
  \bibinfo{person}{Sarah~Charlotte Johnson}, \bibinfo{person}{Kareha~M Agesa},
  \bibinfo{person}{Katya~Anne Shackelford}, \bibinfo{person}{Derrick Tsoi},
  \bibinfo{person}{Daniel~Rhodes Kievlan}, \bibinfo{person}{Danny~V Colombara},
  \bibinfo{person}{Kevin~S Ikuta}, \bibinfo{person}{Niranjan Kissoon},
  \bibinfo{person}{Simon Finfer}, {et~al\mbox{.}}}
  \bibinfo{year}{2020}\natexlab{}.
\newblock \showarticletitle{Global, regional, and national sepsis incidence and
  mortality, 1990--2017: analysis for the Global Burden of Disease Study}.
\newblock \bibinfo{journal}{\emph{The Lancet}} \bibinfo{volume}{395},
  \bibinfo{number}{10219} (\bibinfo{year}{2020}), \bibinfo{pages}{200--211}.
\newblock


\bibitem[Shashikumar et~al\mbox{.}(2017)]%
        {shashikumar2017early}
\bibfield{author}{\bibinfo{person}{Supreeth~P Shashikumar},
  \bibinfo{person}{Matthew~D Stanley}, \bibinfo{person}{Ismail Sadiq},
  \bibinfo{person}{Qiao Li}, \bibinfo{person}{Andre Holder},
  \bibinfo{person}{Gari~D Clifford}, {and} \bibinfo{person}{Shamim Nemati}.}
  \bibinfo{year}{2017}\natexlab{}.
\newblock \showarticletitle{Early sepsis detection in critical care patients
  using multiscale blood pressure and heart rate dynamics}.
\newblock \bibinfo{journal}{\emph{Journal of electrocardiology}}
  \bibinfo{volume}{50}, \bibinfo{number}{6} (\bibinfo{year}{2017}),
  \bibinfo{pages}{739--743}.
\newblock


\bibitem[Singer et~al\mbox{.}(2016)]%
        {10.1001/jama.2016.0287}
\bibfield{author}{\bibinfo{person}{Mervyn Singer}, \bibinfo{person}{Clifford~S.
  Deutschman}, \bibinfo{person}{Christopher~Warren Seymour},
  \bibinfo{person}{Manu Shankar-Hari}, \bibinfo{person}{Djillali Annane},
  \bibinfo{person}{Michael Bauer}, \bibinfo{person}{Rinaldo Bellomo},
  \bibinfo{person}{Gordon~R. Bernard}, \bibinfo{person}{Jean-Daniel Chiche},
  \bibinfo{person}{Craig~M. Coopersmith}, \bibinfo{person}{Richard~S.
  Hotchkiss}, \bibinfo{person}{Mitchell~M. Levy}, \bibinfo{person}{John~C.
  Marshall}, \bibinfo{person}{Greg~S. Martin}, \bibinfo{person}{Steven~M.
  Opal}, \bibinfo{person}{Gordon~D. Rubenfeld}, \bibinfo{person}{Tom van~der
  Poll}, \bibinfo{person}{Jean-Louis Vincent}, {and} \bibinfo{person}{Derek~C.
  Angus}.} \bibinfo{year}{2016}\natexlab{}.
\newblock \showarticletitle{{The Third International Consensus Definitions for
  Sepsis and Septic Shock (Sepsis-3)}}.
\newblock \bibinfo{journal}{\emph{JAMA}} \bibinfo{volume}{315},
  \bibinfo{number}{8} (\bibinfo{date}{02} \bibinfo{year}{2016}),
  \bibinfo{pages}{801--810}.
\newblock
\showISSN{0098-7484}
\urldef\tempurl%
\url{https://doi.org/10.1001/jama.2016.0287}
\showDOI{\tempurl}


\bibitem[Tzeng et~al\mbox{.}(2014)]%
        {tzeng2014deep}
\bibfield{author}{\bibinfo{person}{Eric Tzeng}, \bibinfo{person}{Judy Hoffman},
  \bibinfo{person}{Ning Zhang}, \bibinfo{person}{Kate Saenko}, {and}
  \bibinfo{person}{Trevor Darrell}.} \bibinfo{year}{2014}\natexlab{}.
\newblock \showarticletitle{Deep domain confusion: Maximizing for domain
  invariance}.
\newblock \bibinfo{journal}{\emph{arXiv preprint arXiv:1412.3474}}
  (\bibinfo{year}{2014}).
\newblock


\bibitem[van Wyk et~al\mbox{.}(2019)]%
        {8624374}
\bibfield{author}{\bibinfo{person}{Franco van Wyk}, \bibinfo{person}{Anahita
  Khojandi}, {and} \bibinfo{person}{Rishikesan Kamaleswaran}.}
  \bibinfo{year}{2019}\natexlab{}.
\newblock \showarticletitle{Improving Prediction Performance Using Hierarchical
  Analysis of Real-Time Data: A Sepsis Case Study}.
\newblock \bibinfo{journal}{\emph{IEEE Journal of Biomedical and Health
  Informatics}} \bibinfo{volume}{23}, \bibinfo{number}{3}
  (\bibinfo{year}{2019}), \bibinfo{pages}{978--986}.
\newblock
\urldef\tempurl%
\url{https://doi.org/10.1109/JBHI.2019.2894570}
\showDOI{\tempurl}


\bibitem[Weimann and Conrad(2021)]%
        {weimann2021transfer}
\bibfield{author}{\bibinfo{person}{Kuba Weimann} {and} \bibinfo{person}{Tim~OF
  Conrad}.} \bibinfo{year}{2021}\natexlab{}.
\newblock \showarticletitle{Transfer learning for ECG classification}.
\newblock \bibinfo{journal}{\emph{Scientific reports}} \bibinfo{volume}{11},
  \bibinfo{number}{1} (\bibinfo{year}{2021}), \bibinfo{pages}{1--12}.
\newblock


\bibitem[Xie et~al\mbox{.}(2020)]%
        {xie2020epidemiology}
\bibfield{author}{\bibinfo{person}{Jianfeng Xie}, \bibinfo{person}{Hongliang
  Wang}, \bibinfo{person}{Yan Kang}, \bibinfo{person}{Lixin Zhou},
  \bibinfo{person}{Zhongmin Liu}, \bibinfo{person}{Bingyu Qin},
  \bibinfo{person}{Xiaochun Ma}, \bibinfo{person}{Xiangyuan Cao},
  \bibinfo{person}{Dechang Chen}, \bibinfo{person}{Weihua Lu}, {et~al\mbox{.}}}
  \bibinfo{year}{2020}\natexlab{}.
\newblock \showarticletitle{The epidemiology of sepsis in Chinese ICUs: a
  national cross-sectional survey}.
\newblock \bibinfo{journal}{\emph{Critical Care Medicine}}
  \bibinfo{volume}{48}, \bibinfo{number}{3} (\bibinfo{year}{2020}),
  \bibinfo{pages}{e209--e218}.
\newblock


\bibitem[Zabihi et~al\mbox{.}(2019)]%
        {zabihi2019sepsis}
\bibfield{author}{\bibinfo{person}{Morteza Zabihi}, \bibinfo{person}{Serkan
  Kiranyaz}, {and} \bibinfo{person}{Moncef Gabbouj}.}
  \bibinfo{year}{2019}\natexlab{}.
\newblock \showarticletitle{Sepsis prediction in intensive care unit using
  ensemble of XGboost models}. In \bibinfo{booktitle}{\emph{2019 Computing in
  Cardiology (CinC)}}. IEEE, \bibinfo{pages}{Page--1}.
\newblock


\bibitem[Zhang and Yang(2021)]%
        {zhang2021survey}
\bibfield{author}{\bibinfo{person}{Yu Zhang} {and} \bibinfo{person}{Qiang
  Yang}.} \bibinfo{year}{2021}\natexlab{}.
\newblock \showarticletitle{A survey on multi-task learning}.
\newblock \bibinfo{journal}{\emph{IEEE Transactions on Knowledge and Data
  Engineering}} (\bibinfo{year}{2021}).
\newblock


\bibitem[Zhu et~al\mbox{.}(2019)]%
        {zhu2019deep}
\bibfield{author}{\bibinfo{person}{Ligeng Zhu}, \bibinfo{person}{Zhijian Liu},
  {and} \bibinfo{person}{Song Han}.} \bibinfo{year}{2019}\natexlab{}.
\newblock \showarticletitle{Deep leakage from gradients}.
\newblock \bibinfo{journal}{\emph{Advances in neural information processing
  systems}}  \bibinfo{volume}{32} (\bibinfo{year}{2019}).
\newblock


\end{thebibliography}

\clearpage

\appendix
\label{sec:appendix}
\onecolumn
\section{Supplement Information}

In this section, we list the feature description in detail in Table \ref{table:features} and the standard of Sequential Organ Failure Assessment (SOFA) in Table \ref{table:sofa} which is given in Sepsis-3 \cite{10.1001/jama.2016.0287}.

\begin{table*}[h]
  \centering
  \caption{Feature Description}
  \label{table:features}
  \begin{tabular}{ll} 
  \toprule
    \textbf{Type} & \textbf{Features}\\  \midrule
    \textit{\textbf{Demographic variables}} & Age, Gender, ICU\_hours \\  \midrule
    \textit{\textbf{Vital sign variables}} &  Heart rates, Temperature, Systolic BP, Mean arterial pressure, Diastolic BP, Respiration rate \\ \midrule
    \textit{\textbf{Laboratory variables}} & \begin{tabular}[c]{@{}l@{}} $FiO_2$, $SaO_2$, pH, AST, BUN, Calcium, Chloride, Creatinine, Glucose, Potassium, Total Bilirubin, Hct, Hgb, \\ PTT,  WBC, Platelets, BUN/CR,  $SaO_2$/$FiO_2$ \end{tabular} \\ \bottomrule
 \end{tabular}
 \vspace{+1em}
\end{table*}

\begin{table*}[h]
\caption{Sequential {[}Sepsis-Related{]} Organ Failure Assessment Score \cite{10.1001/jama.2016.0287}}
\label{table:sofa}
\small
\begin{tabular}{llllll}
\toprule
\multirow{2}{*}{\textbf{System}} &
  \multicolumn{5}{c}{\textbf{Score}} \\
  \cmidrule(r){2-6}
 &
  \multicolumn{1}{l}{\textbf{0}} &
  \multicolumn{1}{l}{\textbf{1}} &
  \multicolumn{1}{l}{\textbf{2}} &
  \multicolumn{1}{l}{\textbf{3}} &
  \multicolumn{1}{l}{\textbf{4}} \\ \hline
\textit{\textbf{Respiration}} &  &   &   &   &   \\
\begin{tabular}[c]{@{}l@{}}$PaO_{2}/FiO_{2}$, \\ mmHg (kPa)\end{tabular} &  >= 400(53.3) &  < 400(52.3) &  < 300(40) &
  \begin{tabular}[c]{@{}l@{}}<200(26.7) with\\ respiratory support\end{tabular} &
  \begin{tabular}[c]{@{}l@{}}<100 (13.3) with\\ respiratory support\end{tabular} \\ \midrule
\textit{\textbf{Coagulation}} &   &   &  &   &   \\
Platelets, $\times 10^3 /\mu L$ &  >=150 &  < 150 &  < 100 &  <50 &  <20 \\ \midrule
\textit{\textbf{Liver}} &   &   &   &   &   \\
\begin{tabular}[c]{@{}l@{}}Total Bilirubin, mg/dL\\ ($\mu mol/L$)\end{tabular} &
  <1.2(20) &  1.2-1.9 (20-32) &  2.0-5.9 (33-101) & 6.0-11.9 (102-204) &  > 12.0 (204) \\ \midrule
\textit{\textbf{Cardiovascular}} &  MAP >= 70 mmHg &  MAP \textless 70 mmHg &
  \begin{tabular}[c]{@{}l@{}}Dopamine < 5 or \\ dobutamine (any dose)\end{tabular} &
  \begin{tabular}[c]{@{}l@{}}Dopamine 5.1-15\\ or epinephrine <= 0.1\\ or norepinephrine <= 0.1\end{tabular} &
  \begin{tabular}[c]{@{}l@{}}Dopamine > 15 or\\ epinephrine > 0.1\\ or norepinephrine > 0.1\end{tabular} \\ \midrule
\textit{\textbf{Central nervous system}} &   &   &   &   &   \\
Glasgow Coma Scale score &   15 &  13-14 &  10-12 &  6-9 &  <6 \\ \midrule
\textit{\textbf{Renal}} &   &   &   &   &   \\
Creatinine, mg/dL ($\mu mol/L$) &  < 1.2 (110) &  1.2-1.9 (110-170) &  2.0-3.4 (171-299) &  3.5-4.9 (300-440) &  >5.0 (440) \\
Urine output, mL/d &  &   &   &  < 500 &  < 200 \\ \bottomrule
\end{tabular}
\end{table*}

\section{Experiment Environment}
The experiment is conducted on a server with AMD Ryzen 9 3900X 12-Core Processor, 64 GB RAM and GeForce RTX 3090. 
The code is implemented based on Pytorch 1.8.0. 
To train the model, we use Adam \cite{kingma2014adam} with the batch size of 32, and the learning rate is set to 1e-3. 
We repeat each experiment for 5 times (i.e., 5 seeds) and record the average results.

\end{document}